# A Survey on Deep Learning Methods for Robot Vision

Javier Ruiz-del-Solar, Patricio Loncomilla, Naiomi Soto

Advanced Mining Technology Center & Dept. of Electrical Engineering
Universidad de Chile, Chile
{jruizd@ing.uchile.cl, ploncomi@ing.uchile.cl, c.naiomi.soto@gmail.com}

**Abstract.** Deep learning has allowed a paradigm shift in pattern recognition, from using hand-crafted features together with statistical classifiers to using general-purpose learning procedures for learning data-driven representations, features, and classifiers together. The application of this new paradigm has been particularly successful in computer vision, in which the development of deep learning methods for vision applications has become a hot research topic. Given that deep learning has already attracted the attention of the robot vision community, the main purpose of this survey is to address the use of deep learning in robot vision. To achieve this, a comprehensive overview of deep learning and its usage in computer vision is given, that includes a description of the most frequently used neural models and their main application areas. Then, the standard methodology and tools used for designing deep-learning based vision systems are presented. Afterwards, a review of the principal work using deep learning in robot vision is presented, as well as current and future trends related to the use of deep learning in robotics. This survey is intended to be a guide for the developers of robot vision systems.

## 1 Introduction

Deep learning[1] is a hot topic in the pattern recognition, machine learning, and computer vision research communities. This can be seen clearly in the large number of reviews, surveys, special issues, and workshops that are being presented, and the special sessions in conferences that address this topic (e.g. [9][10][11][12][13][14][16][55][56][57]). Indeed, the explosive growth of computational power and training datasets, as well as technical improvements in the training of neural networks, has allowed a paradigm shift in pattern recognition, from using hand-crafted features (e.g. HOG, SIFT, and LBP) together with statistical classifiers to the use of data-driven representations, in which features and classifiers are learned together. Thus, the success of this new paradigm is that "it requires very little engineering by hand" [12], because most of the parameters are learned from the data, using general-purpose learning procedures. Moreover, the existence of public repositories with source code and parameters of trained deep neural networks, as well as the existence of specific deep learning frameworks/tools such as *Caffe* [60], has promoted increasing the use of deep learning methods.

The use of the deep learning paradigm has facilitated addressing several computer vision problems in a more successful way than with traditional approaches. In fact, in several computer vision benchmarks, such as the ones addressing image classification, object detection and recognition, semantic segmentation, and action recognition, just to name a few, most of the competitive methods are now based on the use of deep learning techniques (see Table 2). In addition, most of the recent presentations at the flagship conferences in this area (e.g. CVPR, ECCV, ICCV) use deep learning methods or hybrid approaches that incorporate deep learning.

---

[1] The term *deep* is used to denote artificial neural networks with several layers, as opposed to traditional artificial neural networks with few layers that are therefore called *shallow* networks.

Deep learning has already attracted the attention of the robot vision community [56][57]. However, given that new methods and algorithms are usually developed within the computer vision community and then transferred to the robot vision community, the question is whether or not new deep learning solutions to computer vision and recognition problems can be directly transferred to robot vision applications. We believe that this transfer is not straightforward considering the multiple requirements of current deep learning solutions in terms of memory and computational resources, which in many cases include the use of GPUs. Furthermore, following [54], we believe that this transfer must consider that robot vision applications have different requirements from standard computer vision applications, such as real-time operation with limited on-board computational resources, and the constraining observational conditions derived from the robot geometry, limited camera resolution, and sensor/object relative pose.

Currently, there are several reviews and surveys related to deep learning [9][10][11][12][13][14][16]. However, they are not focused on robotic vision applications, and they do not consider their specific requirements. In this context, the main motivation of our survey is to address the use of deep learning in robot vision. In order to achieve this, first, an overview of deep learning and its use in computer vision is given, which includes a description of the most frequently used neural models and their main application areas. Then, the standard methodology and tools used for designing deep learning-based vision systems are presented. Afterwards, a review of the main current work using deep learning in robot vision is presented, as well as current and future trends related to the use of deep learning in robotics.

This survey is intended to be a guide for developers of robot vision systems; therefore the focus is on the practical aspects of the use of deep neural networks rather than on theoretical issues. It is also important to note that as Convolutional Neural Networks [16][17] are the most commonly used deep neural networks in vision applications, we focus our analysis on them.

This paper is organized as follows: in Section 2, the use of deep neural networks in computer vision is described. The historical development of the discipline is presented first; then some basic definitions are given; and finally the most important deep neural models are described, as well as their applications in computer vision. In Section 3, the design process of deep-learning based vision applications, including the case of robot vision systems, is discussed. In Section 4, a review of the recent application of deep learning in robot vision systems is presented. In Section 5 current and future challenges of deep learning in robotic applications are discussed. Finally, in Section 6, conclusions related to the use of deep learning in robotics are drawn.

The reader interested in having a practical guide for the use/application of deep learning in robot vision, and not in the basic aspects of deep learning, is referred to Sections 3-6.

## 2 Deep Neural Networks in Computer Vision

Deep learning is based on the use of artificial neural networks (ANNs) with several hidden layers. The basic aspects of deep neural networks (DNNs) and their usage in computer vision applications will be described in this section. In Section 2.1, we will review the historical development of DNNs, and in Section 2.2, we will present some basic definitions needed to understand the different DNN models. Then, in Sections 2.3-2.6, we will present various DNN models used in computer vision applications, and in Section 2.7 the Deep Reinforcement Learning paradigm. Finally, in Section 2.8, we will present techniques used in order to reduce the memory and computational requirements of DNNs.

## 2.1. Historical Development

ANNs were first proposed in 1943 as electrical circuits by Warren McCulloch, and a learning algorithm was proposed by Donald Hebb in 1949 [185]. Marvin Minsky built the first ANN in hardware in 1951, and Frank Rosenblatt in 1958 built the first ANN that was able to learn. These ANNs were shallow networks, i.e., typically, they were formed by three or four layers of neurons, with only one or two hidden layers. In [186], Poggio and Girosi demonstrated that a regularized learning algorithm was equivalent to an interpolation using one Radial Basis Function (RBF) per each datum in the input, which can be reinterpreted as an ANN. Funahashi [187] was able to show that a three-layer, shallow neural network is able to represent any continuous function with arbitrary precision. This result can be extended to networks with more layers in a straightforward way, by induction. However, this work was not able to guarantee that a training procedure can achieve the theoretical optimal configuration.

In 1980, Fukushima proposed the NeoCognitron [4], a multilayered neural network inspired by the seminal studies of the mammalian visual system of Hubel and Wiesel [3]. The NeoCognitron is composed of a sequence of two-dimensional layers of neurons, with each neuron being connected to near neurons from the previous layer. Each layer can be formed by "S-cells" (extract visual features), "C-cells" (pool local responses), or "V-cells" (average values). Similar 2D neuron layers are grouped into 3D cell planes. The NeoCognitron has no supervised training algorithm; it has to be trained by specifying the type of expected response for each layer separately, i.e., the weights for the neurons are not computed in a fully automatic way.

Inspired by the work of Fukushima, LeCun et al. [17][18][19] developed the first Convolutional Neural Network (CNN) in the late 1980s. It included the concept of feature maps produced by layers of neurons with local and heavily restricted connections [17] (similar to the receptive fields of biological neurons), and whose weights were adapted using the backpropagation algorithm [70]. Although the network was successful in the recognition of handwritten characters, its further development was delayed by the limited computational capabilities available at that time, and by some algorithmic difficulties when training deep networks. In particular, gradient vanishing [1][6] in deep networks was one of the most important issues. Also, the result that a shallow network can approximate any function [187] and the lack of strong theoretical warranties of global convergence [12] when compared to alternative methods like SVM [2] were important factors that hindered the acceptance of deep learning methods within the machine learning community. Thus, the use of handcrafted features used together with statistical classifiers (sometimes implemented as shallow networks) continued to be the dominant paradigm in pattern recognition and machine learning.

During the 1990s and the beginning of the 21st Century, deep feedforward networks (e.g. CNN) were largely forsaken by the machine-learning community and ignored by the computer vision community [12] (e.g. a 2002 review on image processing with neural networks [61] makes almost no mention of DNNs). However, interest in these networks was reborn in mid-2000 (e.g. [71][72][73]). But it was not until the proposal of the AlexNet [20] that the boom of deep neural networks started in the computer vision community. AlexNet was able to largely outperform previous feature-based methods in the ILSVRC 2012 challenge (AlexNet brought down the error rate by about 40% compared to hand-engineering approaches), a well-known image classification benchmark. The main factors explaining the success of AlexNet are the use of specific technical solutions such as ReLU activation functions, local response normalization, dropout and stochastic gradient descent, in addition to the use of GPUs for training and testing the network, using large amounts of data (the ImageNet dataset [69] in the case of AlexNet). All these factors have enabled CNNs to be in the core of the most current state-of-the-art applications for computer vision.

Table 1 shows a summary of relevant work in deep learning for computer vision. All the models and algorithms mentioned will be described in the following sections.

Table 1. Summary of Relevant Work in Deep Learning for Computer Vision

| Paper | Contribution | Model/Algorithm |
| --- | --- | --- |
| Fukushima, 1980 [4] | The Neocognitron network is proposed. | Neocognitron |
| LeCun et al., 1998 [19] | LeNet-5, the first widely known convolutional neural network, is proposed. Previous versions of this network were proposed in 1989-1990. [17][18]. | LeNet-5 |
| Hochreiter & Schmidhuber, 1997 [68] | LSTM recurrent networks are introduced. | LSTM (Long Short-Term Memory) |
| Nair & Hinton, 2010 [5] | The paper introduced the ReLU activation functions. | ReLU (Rectified linear unit) |
| Glorot, Bordes, & Bengio, 2010 [7] | The paper demonstrated that the training of a network is much faster when ReLU activation functions are used. | ReLU (Rectified linear unit) |
| Bottou, 2010 [67] | The Stochastic Gradient Descent is proposed as a learning algorithm for large-scale networks. | Stochastic Gradient Descent |
| Hinton et al. 2012 [62] | Dropout, a simple and effective method to prevent overfitting, is proposed. | Dropout |
| Krizhevsky et al., 2012 [20] | AlexNet is proposed and, thanks to its outstanding performance in the ILSVRC 2012 benchmark [69], the boom of deep learning in the computer vision community started. AlexNet has 8 layers. | AlexNet |
| Simonyan & Zisserman, 2014 [21] | The VGG network is proposed. It has 16/18 layers. | VGG |
| Girshick et al., 2014 [27] | The Regions with CNN features network is proposed. | R-CNN |
| Szegedy et al., 2015 [22] | The GoogleNet network is proposed. It has 22 layers. | GoogleNet |
| He et al., 2015 [23][24] | The ResNet network is proposed. It is based on the use of residuals and is able to use up to 1001 layers. | ResNet |
| Badrinarayanan et al., 2015 [31] | SegNet, a fully convolutional network for image segmentation applications, is proposed. | SegNet |

| Van de Sande et al., 2016 [235] | The DenseNet network is proposed. It includes skip connections between every layer and its previous layers. | DenseNet |
|---|---|---|
| Hu et al [221] | The Squeeze-and-excitation network is proposed. It introduces squeeze-and-excitation blocks used for representing contextual information. | SENet |

## 2.2. Basic Definitions

An artificial neuron takes on an input vector and outputs a scalar value. The neuron is parameterized by a set of weights. Each weight is used as a multiplier for a scalar input. The output of the neuron is the result of applying a nonlinear activation function on the sum of the weighted inputs. Thus, a neuron with weights $w$, inputs $x$, output $y$, and non-linear activation function $\phi$ is represented as:

$$y = \phi\left(\sum_{i=0}^{n} w_i \cdot x_i\right)$$

with $n$ the size of $x$, and $w_0$ representing a bias for the neuron input if $x_0$ is set to 1.

Non-linear activation functions $\phi(x)$ are used for improving the expressive power of the network. They are valid as long as they are continuous, bounded, and monotonically increasing [65]. An additional requirement of the learning algorithms over $\phi(x)$ is differentiability. Thus, the activation functions used most often include the sigmoid logistic function $\phi(x) = \frac{1}{1+\exp(x)}$, the sigmoid hyperbolic tangent function $\phi(x) = \tanh(x)$, the rectified linear unit (ReLU) function $\phi(x) = \max(0, x)$ [5], and the parametric rectified linear unit (PReLU) $\phi(x) = \max(0, x) + \alpha \min(0, x)$ [8]. In addition, activation functions that output a vector like *softmax* $\phi_j(x) = \frac{\exp(x_j)}{\sum_{k=1}^{m} \exp(x_k)}$, whose outputs add up to 1 and enable them to represent multi-class output probabilities [5], are selected for neurons used as multi-class classifiers.

An artificial neural network (ANN) consists of a set of connected neurons. Typically, neurons are grouped in layers. Each layer takes a set of inputs for computing a set of outputs. The input-output relation is determined by the weights in the layer. A shallow network contains a small number of hidden neuron layers, and one output layer. Shallow feedforward networks are able to represent any function with an arbitrary number of hidden neurons [65]. However, training processes are not guaranteed to achieve the global optimum. DNNs are artificial neural networks containing several hidden layers. They are also able to represent any function with an arbitrarily high number of hidden neurons. Moreover, under certain conditions, feedforward deep networks are able to learn functions using fewer neurons than shallow networks [77][78].

A DNN, and in general any ANN, is specified by both hyper-parameters and parameters. Hyper-parameters are related to the architecture of the network (number of layers, number of neurons per layer, activation function, connectivity of the neurons, etc.)

and to the parameters of the learning process. The parameters of a network correspond to the set of weights of all the neurons. The network can be trained for selecting optimal weights (parameters) by using numerical optimization methods. The optimal hyper-parameters cannot be learned directly from the data, so several network architectures must be trained and tested for selecting the best hyper-parameters. Thus, in order to be able to adapt the parameters of the network, to select its hyper-parameters, and to characterize its performance, the available data must be divided into three non-overlapping datasets: training, validation, and test. The training dataset is used for learning the model parameters (weights) of the network. The validation dataset is used for selecting the hyper-parameters by evaluating the performance of the network under different hyper-parameter configurations. The test dataset is used for characterizing the trained network by estimating its generalization error.

DNNs are usually trained using supervised learning. The objective of the training procedure is to find the network parameters that minimize a loss function (e.g. a function of the global error between the obtained and the desired output of the network). Two kinds of iterative optimization procedures are commonly used: first order optimization (gradient based), and second order optimization (hessian based). When the loss function is convex, second order optimization is commonly used because it converges faster to the optimal solution. However, in non-convex problems, second order algorithms can be attracted to saddle points or local minima [79][80]. Then, gradient-based methods are normally used for training deep neural networks. As the datasets used for training deep neural networks are generally large, only a small subset of the data (a mini-batch) is used at each optimization iteration for training the network. This procedure is called Stochastic Gradient Descent (SGD) [67]. The term stochastic is used because each mini-batch gives a noisy estimation of the gradient over all the data, which allows finding a good set of weights very quickly [12]. "The stochastic gradient descent directly optimizes the expected risk, since the examples are randomly drawn from the ground truth distribution." [67]. The use of mini-batches improves the generalization ability of the network and enables fast training. Additional mechanisms are the decrease of the learning rate over time, and the use of momentum [66]. The generation of specialized second order algorithms for deep learning is a current research topic [79].

In several cases, minimizing the original loss function can cause misbehavior of the optimization process. For example, in some cases the learned weights can be too large because of overfitting. When the training procedure generates ill-posed networks, regularization can be used. Regularization corresponds to a set of techniques used to reduce overfitting by improving robustness in the parameter estimation process [80]. Some of these techniques can be implemented by adding a regularization term to the loss function. This extra term can be used for limiting the size of the weights, and also for improving invariance against small changes in the inputs. As an example, L2 regularization [80] consists of adding a penalization term to the original loss function:
$L_{regul}(x,w) = L_{original}(x,w) + \alpha * \|w\|^2$.

Dropout [62][63] is a technique that consists of using only a random subset of the neurons in each layer (along with its connections) during training episodes in order to reduce overfitting. This procedure is able to improve the generalization error of the networks, since its effect is similar to that of using an ensemble of networks. Using dropout can be interpreted as an alternative form of regularization. Once the network has been trained, weights are decreased proportionally in order to account for the smaller number of neurons used when training. Also, dropout can be used for generating multiple possible outputs for the same input [32] when activating different units. In this way, a confidence level for the output can be computed. An alternative to Dropout is DropConnect [64], which, instead of dropping the activations of some neurons, drops their weights. In this way each unit receives input from a random subset of units on the previous layer.

Batch normalization consists of normalizing the input layers at each mini-batch iteration [177]. This enables the training procedure to use higher learning rates, and to be more robust with respect to initial weights in the network. Batch normalization also acts as a regularizer, eliminating the need for dropout. State-of-the-art ResNet networks, described in Section 2.3, do not use dropout; they use batch normalization [23].

## 2.3 Convolutional Neural Networks

Convolutional Neural Networks (CNNs) correspond to a family of DNN architectures whose main goal is analyzing natural signals by taking advantage of their properties. According to [12], the main properties of CNNs are local connections, shared weights, pooling, and the use of many layers. The main reason for this is that the structure of CNNs is inspired by the processing pipeline of the mammalian visual system [3], where (i) neurons have local connectivity, (ii) several layers of neurons exist between the fovea and the cortical visual areas, (iii) neurons on a given layer perform the same kind of processing (a non-linear activation function is applied over its weighted inputs), and (iv) the size of the receptive fields of the neurons (i.e. the region of the sensory space whose signals affect the operation of a given neuron) increases from one layer to the next one. Hence, the behavior of a given layer of CNN neurons can be modeled as a linear filtering operation, i.e. a convolution, followed by a non-linear operation (activation function). After the convolution layer, the application of a pooling operation permits modeling the increase of the receptive size of the neurons. After several convolutions and pooling operations are applied (the network has several layers), a decision needs to be made about the input stimuli. This is implemented using fully connected layers of neurons. Thus, CNNs are composed of three kinds of processing layers: convolutional layers, pooling layers, and fully-connected layers, as is shown in Figure 1.

The number of layers in a CNN and their type will depend on the specific network structure, but normally the number of convolution and pooling layers is much larger than the number of fully-connected layers. A very important side effect of this is that a CNN can have a smaller number of weights to be adapted than a shallow network, given that the number of free parameters in the convolutional layers is very small, and proportional to the size of the implemented filters. For instance, in a CNN whose first convolutional layer has 224x224x3 input neurons (i.e., 3 channels), a filter size of 11x11, and C output channels, the number of free parameters, i.e. the weights, to be adapted is 11x11x3xC, while in the case of a fully-connected network, the number of weights would be 224*224*3*C. Then, the number of free parameters increases linearly if the input has several channels (i.e., a RGB input image) and/or if several output channels must be computed. One important aspect to highlight is that most of the operations that need to be computed in a CNN are local, and therefore can be mapped easily to GPUs.

The designing of a convolutional layer basically consists of selecting the number of filters that a given layer will have, and its size. The parameters of the filters, i.e. the weights of the neurons, will be adapted during training. In the case of the design of a pooling layer, the pooling operation needs to be selected. The most frequently used pooling operations are average pooling and max pooling; however, several other alternatives have been studied and used [9] (e.g. stochastic pooling, spatial pyramid pooling, def-pooling).

The fully-connected layers are the ones in charge of converting the 2D feature maps into a 1D feature vector, in addition to being in charge of the classification. The fully-connected layers have a similar structure to a traditional shallow network, and they normally contain a large percentage of the parameters of a CNN (about 90% according to [9]). In some cases the fully-connected layers in charge of the final classification are replaced by a *softmax* unit, or a different statistical classifier (e.g. an SVM [127]). *Softmax* has been shown to achieve better accuracy than SVM [28], because it can be trained end-to-end inside the network.

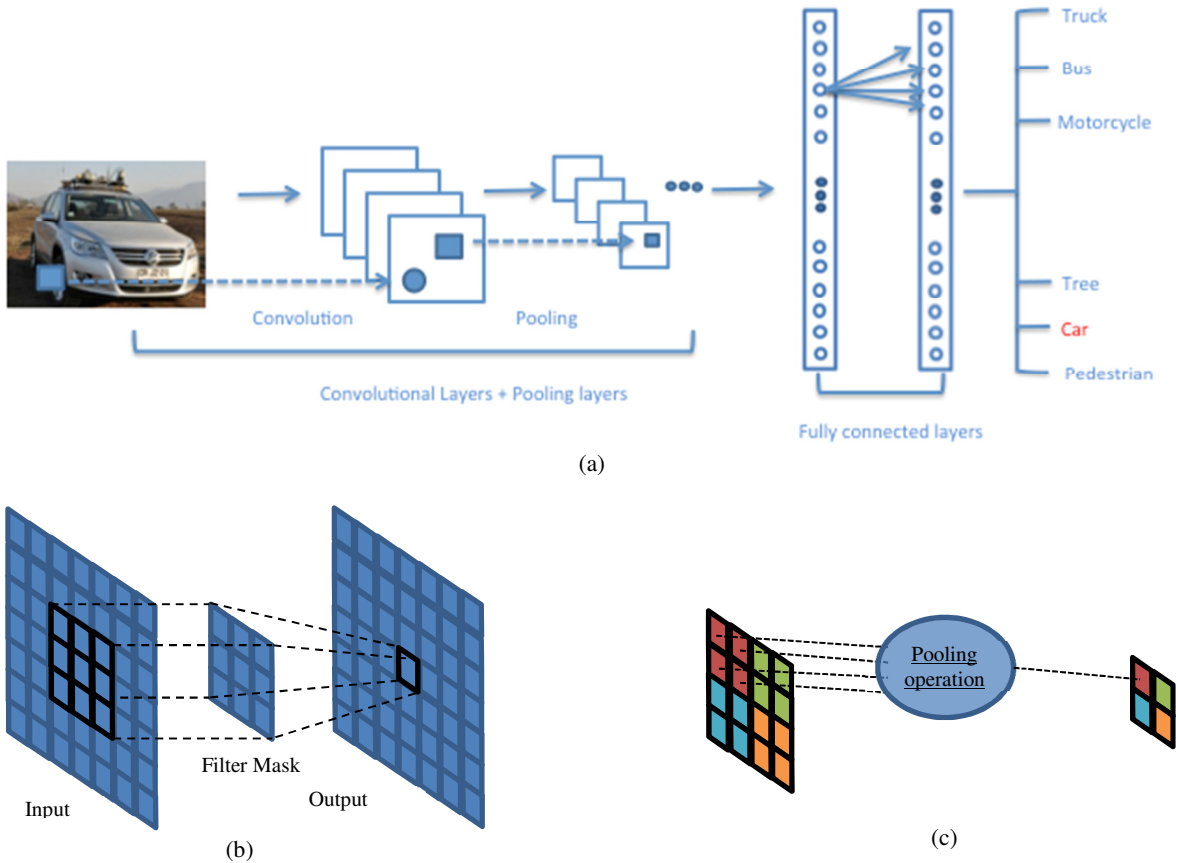

Figure 1. (a) CNN Architecture. (b) The Operation of a Convolutional layer. (c) The Operation of a Pooling/Resampling layer. Adapted from [9].

Different CNN architectures implement different network topologies (e.g. number of convolutional layers, filter size in each layer, number of fully connected layers), and operations (e.g. activation function, pooling operation, dropout). AlexNet, one of the most popular CNNs and the one responsible for the DNN boom thanks to its outstanding performance in ILSVRC2012, has 5 convolutional layers and 3 fully connected layers. The convolutional layers 1, 2, and 5 include max pooling, and the convolutional layers 1 and 2 include local response normalization [20]. In addition, they use ReLU activation functions, dropout, and stochastic gradient descent for training. AlexNet has a fixed input size (224x224x3) and is well suited for image classification tasks. The network generates a probability vector of size 1000 from the whole input image, each component representing the probability of presence for one object category.

Some CNN architectures keep the same working strategy as AlexNet, but they go deeper. For instance, VGG [21], which got the second place at ILSVRC 2014, has 16 (18) layers, 13 (15) convolutional and 3 fully-connected, but it reduces the filter size to 3x3. GoogleNet [22], which won first place at ILSVRC 2014, has 22 layers, and introduced a more complex architecture based on the use of several parallel sub-layers inside each layer. However, in practice, the use of 30 or more layers generates an increase in the error. This problem is solved by ResNet [23] by using residual layers that work by adding the output of a previous layer to the current one. Then, the network processes residuals instead of the original data layers. ResNet is able to use up to 1001 layers [24], consistently decreasing the error when adding more layers. While ResNet uses only skip connections between pairs of layers, DenseNet [235] uses skip connections between every layer and its previous layers. DenseNet has been shown to perform better than ResNet, while requiring fewer parameters to learn. Finally squeeze-and-excitation networks, (SENet) ,[221] were introduced in 2017. As each layer has a reduced local receptive field, channel descriptors are computed for exploiting contextual information outside of this region. SENet was the winner of the ILSVRC 2017 classification challenge, and is currently the state-of-the-art for image classification.

On the other hand, some approaches focus on using multiple CNNs that work independently, and whose results are combined. One of the most interesting approaches is the Multi-Column DNN (MCDNN) [93], which uses several DNNs in parallel and averages their predictions. MCDNN-based approaches obtain state-of-the-art results on some image classification benchmarks (see Table 2).

Table 2. Performance of the Best Performing DNNs in some Selected Computer Vision Benchmarks. The absolute ranking corresponds to the ranking of the network in the benchmark, while the relative one corresponds to the ranking when considering only DNN models.

| Task | Benchmark | Paper | Relative (Absolute) Ranking | DNN Model |
|---|---|---|---|---|
| Image Classification | MNIST | Wan et al., 2013 [173] | 1(1) | MCDNN+ DropConnect |
| Image Classification | CIFAR-10 | Graham, 2015 [174] | 1(1) | DeepCNet (Adapted MCDNN) + Fractional Max Pooling |
| Image Classification | CIFAR-100 | Clevert et al., 2016 [175] | 1(1) | ELU-Network |
| Image Classification | ILSVRC 2017 | Hu et al [221] | 1(1) | Squeeze-and-Excitation networks |
| Object Detection and Recognition | Pascal VOC2012 comp 4 (20 classes) | Dai et al, 2016 [218] | 2(2) | Region-based fully convolutional network |
| Object Detection and Recognition | KITTI (October 11, 2017) (Car/Pedestrian/Cycle) | Ren et al., 2017 [219] | 7/3/3[2](7/3/3) | Recurrent Rolling Convolution |
| Object Detection and Recognition | COCO 2016 (91 Object Types) | Huang et al., 2017 [220] | 1(1) | Faster R-CNN + ResNet |
| Object Detection and Recognition | ILSVRC 2017 Task1a | Shuai et al, 2017 [23][222][223][224][225][226][227] | 1(1) | Residual Attention Network + Feature Pyramid Networks |
| Semantic Segmentation | Cityscapes | DeepLabv3 2017 [208] | 1[2](1) | DeepLab v3 |
| Semantic Segmentation | Pascal VOC2012 comp6 | DeepLabv3-JFT 2017 [208] | 1(1) | DeepLab v3 |

---

[2] The methods obtaining the best performance in this benchmark are not described. It is unknown whether they are based on deep learning or not.

| Action Recognition | MPII | Chen et al. 2017 [228] | 1(1) | Adversarial PoseNet |
| Action Recognition | Pascal VOC2012 comp10 | Gkioxari et al., 2016 [168] | 2(2) | R*CNN |

As mentioned, standard CNN architectures generate outputs for the whole input image. However, in object detection/recognition tasks, where the object of interest covers a small area in the input image, the CNN will not be able to detect the object. In that case, sliding windows can be used for defining the areas of the image to be analyzed. However, this approach, which requires analyzing all possible sliding windows using a CNN, is unable to run in near real-time because of the large number of windows to be processed. Hence, instead of using a sliding window before applying CNNs, region proposals can be computed. A region proposal is a window inside the input image that is likely to contain some object. Region proposal algorithms are class-agnostic object detectors that can be computed very quickly.

The R-CNN algorithm [27] is based on using region proposals that are fed into a CNN. The proposals are computed by using the Selective Search algorithm [26]. The R-CNN algorithm is able to run faster than a naïve CNN using sliding windows, but the CNN features need to be recomputed for each object proposal instance. Fast R-CNN [28] also employs object proposals from Selective Search, but the features to be used for the final detection are computed just once, using shared convolutional layers for each input image. Each proposal defines a region of interest in the last feature map, and the selected region is used for carrying out the classification, using fully connected layers. Faster R-CNN [29] implements a Region Proposal Network (RPN) that is computed by applying a per-pixel fully connected network over the last shared convolutional layer (see Figure 2). The RPN network is able to generate region proposals, which are then used by a Fast R-CNN for the detection of the objects. The Faster R-CNN has also been used with the ResNet architecture [23][24], and it is the state-of-the-art for general-purpose object detection (see Table 2).

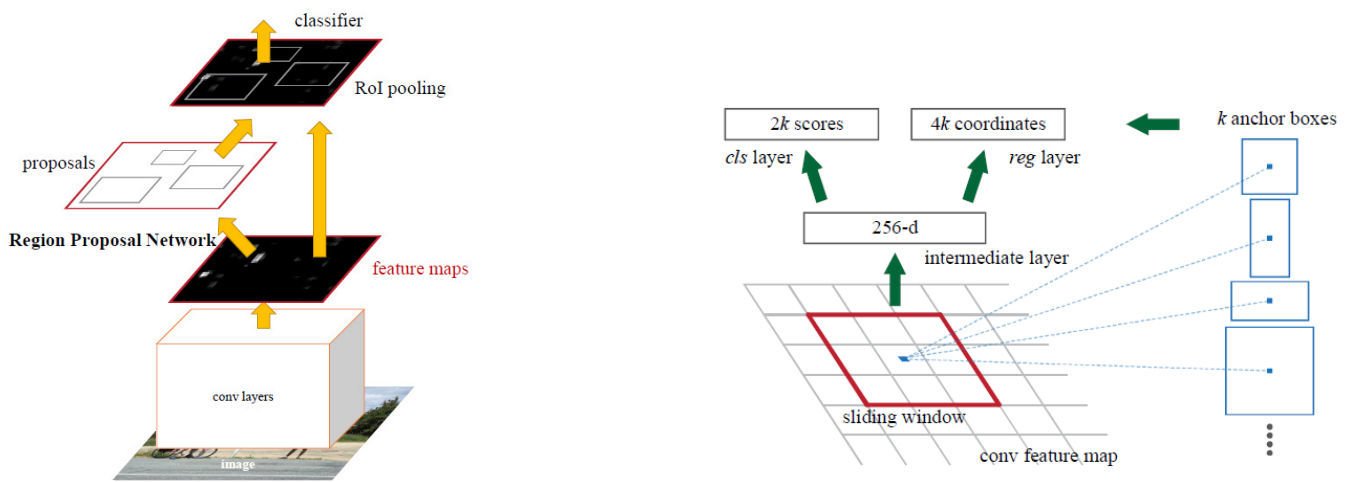

Figure 2: Left: Faster R-CNN is based on computing a set of shared convolutional layers that are used for computing region proposals and then for clasifying them. Right: Region proposal network (RPN). Figures taken from [29] with permission of the authors.

## 2.4 Fully Convolutional Networks

Fully convolutional networks (FCNs) are a special kind of CNN that includes only convolutional layers (no fully connected units). The output of an FCN is a feature map from the last convolutional layer. Given that FCN inputs and outputs are two-dimensional, they are trained end-to-end, pixel-to-pixel, and can be used for dimensionality reduction, regression, or semantic segmentation.

Auto-encoders [81] are artificial neural networks trained to predict their own inputs. FCN based auto-encoders are formed by two smaller FCNs: an encoder and a decoder. The encoder is an FCN that decreases the size of the input image through its layers, generating a compressed representation of the image. The decoder is an FCN that maps the compressed representation onto a final image. When the auto-encoder is trained, its encoder is able to compress images, while its decoder can reconstruct them. Denoising auto-encoders [82] are FCN regression models that are able to denoise images. They are trained with noisy images as input, and their correspondent clean images as output. Also, FCNs can be used for performing semantic segmentation [30][31][32], sometimes in conjunction with Conditional Random Fields (CRFs) [83]. Semantic segmentation FCNs are trained to map RGB images onto images containing a pixel-wise semantic segmentation of the environment.

The first successful use of FCNs for semantic segmentation was implemented in the work of Long, Shelhamer and Darrell [30]. They used a set of convolutional layers with decreasing size whose output corresponds to the segmented image with a lower resolution than the original one. They used dropout during the training stage for regularization. SegNet [31] is also based on an encoder and a decoder. The encoder corresponds to a set of convolutional layers with decreasing spatial size. The decoder takes the output data of the encoder as input, and applies a set of convolutional layers that scale up the data to become the same size as the original input image. By using the decoder, the segmentation obtained by SegNet has higher resolution than that of Long et al. [30]. The SegNet architecture is shown in Figure 3. Bayesian SegNet [32] uses dropout both for training and for normal execution. Since dropout selects random subsets of the neurons when using the network, various network outputs can be obtained by applying dropout several times. Then, a set of semantic, segmented images is available. The images can be analyzed for computing the uncertainty associated with the labels of each pixel. Then, if a pixel has the same label in all of the images, the uncertainty is low, but if the pixel has different labels in the images, their uncertainty is high.

FCNs are based on simple pixel-wise CNNs, without any explicit modeling of the image structure. Thus, current research on semantic segmentation is focused on modifying the baseline FCNs for using contextual information. Dilation10 [83] and DeepLabv2-CRF [85] use dilated/up-sampled filters for covering a greater receptive field without using more parameters. LLR-4x [84] uses a multi-scale representation similar to a Laplace pyramid for collecting semantically and spatially resolved data. DeepLabv2-CRF [85] and Adelaide-Context [86] also integrate CRFs into the FCN process.

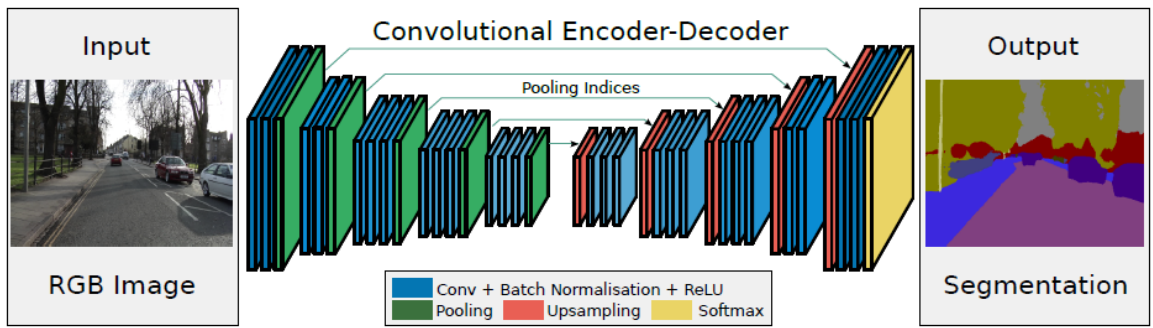

Figure 3: SegNet, an Example of fully Convolutional Architecture. Figure taken from [31] with permission of the authors.

## 2.5 Long Short-Term Memory

Recurrent Neural Networks (RNNs) are networks with at least one feedback connection that allows them to have internal memory. RNNs are useful for learning information from temporal or spatial sequences with variable lengths, which cannot be handled properly by traditional CNNs. A common approach is to self-feed a neural network with its output in the last time frame. Then, back propagation through time (BPTT) [150] can be used for learning the network parameters. However, the use of BPTT suffers from vanishing and exploding gradients [1] because the past network inputs tend to vanish exponentially, or to cause network divergence over time.

A successful recurrent neural network architecture is the Long Short-Term Memory (LSTM) [36][37], which is based on using memory cells. Each memory cell is able to learn simple input-output relations among sequences, even when the relation involves long delays. At each time frame, the memory cell has an input $x_t$, an output $h_t$, and a hidden state $c_t$. The input $x_t$ can modify the hidden state $c_t$ (writing) or be ignored. The output $h_t$ can be related to the hidden state $c_t$ (reading) or be zero. Also, the value of $c_t$ can be preserved in the next frame or set to zero (forget). The amount of reading level $o_t$, writing level $i_t$, and forgetting level $f_t$ at each iteration are computed by using three independent sigmoidal neural layers inside each memory cell. Then, when training the network, the cell learns to store, to output, or to forget values depending on contextual information from the sequences. The architecture of a memory cell does not suffer from exponential vanishing or explosion of the input data. Also, when training the cells by using BPTT, the network state $c_t$ does not require further back propagation.

LSTMs are useful for solving sequence-related problems, such as speech recognition, handwriting recognition, and polyphonic music modeling [128]. By stacking CNN and LSTM based models, systems have been developed that are able to anticipate driver actions [110], generate captions from images [94], answer questions that require image inspection [95], or even align video and text descriptions [96].

## 2.6 Restricted Boltzmann Machines

Restricted Boltzmann Machines (RBMs) are stochastic neural networks composed of stochastic neuron-like binary units. RBMs consist of one layer of visible units, one layer of hidden units, and an extra bias unit. Each visible unit is connected to all hidden units and vice versa. The bias unit is also connected to all hidden and visible units. All of the connections are undirected. RBMs are able to learn the data distribution of the input data (visible units) and to generate samples from that distribution (generative learning) in an unsupervised fashion. The network is able to approximate any distribution as the number of hidden units increases. Given an initial visible input, the network is able to compute binary activations for the hidden units by using binary sampling. Also, given the values of the hidden units, the network can compute activation values for the visible units. By repeating these two processes, the underlying distribution represented by the RBM can be sampled.

The parameters of the network are obtained by minimizing an energy function that depends on the training input vectors (no supervision is required). In practice, parameters are adapted for diminishing the divergence between the distribution sampled from the RBM and the ground truth distribution. Also, several layers of hidden units can be stacked for creating a Deep Boltzmann Machine (DBM), but its training procedure requires high computational resources. Another form of stacking RBMs is by using Deep Belief Networks (DBNs) that include directed connections between layers, enabling the use of greedy layer-wise training algorithms.

A more recent approach for training unsupervised stochastic networks is the Deep Energy Model (DEM) [151]. This architecture is composed of a set of deterministic feed forward layers and a final stochastic layer. This system can also be described as a deep feed-forward network followed by an RBM, trained jointly. Then, the feed-forward network learns a representation that is easily modeled by the final RBM. The system can also be interpreted as a normal RBM with a modified energy function that is dependent on the parameters of the feed-forward network. The training of DEMs is faster than the training of DBMs and DBNs, because only one stochastic layer is used.

DBMs/DBNs/DEMs can be used as a pre-training step for initializing the weights of feed-forward networks. However, although state-of-the-art feed-forward implementations still do not use RBM-derived methods for pre-training [23][80], this tendency could change in the future.

## 2.7 Deep Reinforcement Learning

Reinforcement Learning consists of a set of techniques designed to maximize the cumulative reward obtained by an agent when interacting with its environment. The agent has an internal state, and is able to observe its environment partially, to act on it, and to obtain an immediate reward. This problem can be modeled by using a Markov Decision Process (MDP), which is defined by: (i) $S$, a set of possible states, (ii) $A(s)$, a set of actions possible in state $S$, (iii) $P(s,s',a)$, the probability of transition from $s$ to $s'$ given an action, $a$, (iv) $R(s,s',a)$, the expected reward on transition from $s$ to $s'$ given an action, $a$, and (v) the discount rate, $\gamma$, for delayed reward in discrete time which trades off the importance of immediate and future rewards. Also, a policy, $\pi$, indicates the actions that the agent takes, i.e., it maps each state to an action. The state-action value function, $Q^\pi(s,a)$, of each state-action pair is the expected sum of the discounted rewards for an agent starting at state $s$, taking action $a$ and then following a policy, $\pi$, thereafter. The objective of MDP is to find the optimal policy $\pi^*$ that maximizes the value function. The optimal value function satisfies the Bellman Equation, which has a recursive nature and relates value functions from consecutive time frames. If the optimal state-value function $Q^{\pi^*}(s,a)$ is known, the best next action can be computed trivially without needing to store the policy $\pi^*$ explicitly. When transition probabilities $P(s,s',a)$ and reward function $R(s,s',a)$ are known, the optimal policy can be computed by using dynamic programming. However, when both are not known *a priori,* and can only be estimated by interacting with an environment, other approaches must be used. Classical reinforcement learning, in particular *Q-learning*, uses a discrete state space (usually by partitioning a continuous one) and learns the value function $Q$ by performing several experimental trials, and exploiting the recursivity property for learning from successful ones. Also, approaches that use a linear parameterized function approximation, $Q(s,a;\theta) = \phi(s,a)^T \theta$, with a number of parameters smaller than the number of states, has been used [194]. Deep reinforcement learning (DRL) is able to learn the optimal mapping, $Q^*$, directly from states (represented as sets of consecutive raw observations) and actions related to the different outcomes for each action, by using a convolutional neural network, called a Deep Q Network (DQN). This technique has been applied successfully for agents to learn by playing several Atari games with only raw gray scale images and the scores in the game [190], achieving human-level playing ability. As neural networks learn from new data, a technique named experience replay is used. It consists of selecting random samples from both old and new trials for training the network. Then, the learning process does not get biased by the last trials, and context information from old trials can be preserved and used [190]. Policy gradient methods and actor/critic algorithms have been proposed recently [195] for enabling end-to-end training and faster convergence and evaluation times with respect to DQNs. The use of supervised learning for policy networks as initialization for reinforcement learning has proven to be useful, and has enabled computers to defeat human professional players in the game of Go [196]. Deep reinforcement learning is a very hot research topic as it has the potential to solve real-world problems with a human-level performance, using end-to-end formulations.

## 2.8 Deep Compression and Quantization

State-of-the-art vision systems based on CNNs require large memory and computational resources, such as those provided by high-end GPUs. For this reason CNN-based methods are unable to run on devices with low resources, such as smartphones or mobile robots, limiting their use in real-world applications. Thus, the development of mechanisms that allow CNNs to work using less memory and fewer computational resources, such as compression and quantization of DNNs, is an important area to be addressed.

Mathieu et al. [183] propose computing convolutions by using FFTs. This technique enables reusing transformed filters with multiple images. FFT is efficient when computing responses for shared convolutional layers. Also, inverse FFT is required only once for each output channel. However, the use of FFTs requires additional memory for storing the layers in the frequency domain. Libraries such as CuDNN [153] are able to use FFTs when training and testing the networks. FFT is more efficient than normal convolutions when filter masks are large. However, the state-of-the-art ResNet [23] uses small 3x3 masks.

The use of sparse representations is a useful approach for network compression. In [87], kernels are factorized into a set of column and row filters, by using an approach based on separable convolutions. A 2.5x speedup is obtained with no loss of accuracy, and a 4.5x speedup with a loss in accuracy of only 1%. A procedure for reducing the redundancy in a CNN by using sparse decomposition is proposed in [88]. Both the images and the kernels are transformed onto a different space by using matrix multiplications, where convolutions can be computed by using sparse matrix multiplication. This procedure is able to make more than 90% of the parameters zero in an AlexNet [20], with a drop in accuracy that is less than 1%.

In [89], a methodology for compressing CNNs is proposed, which is based on a three-stage pipeline: pruning, trained quantization, and Huffman coding. In the first stage, the original network is pruned by learning only the important connections, and a compressed sparse row/column format is used for storing the non-zero components. In the second stage, weights are quantized by using a 256-word dictionary for convolutional layers, and a 32-word dictionary for fully connected layers, enabling weight sharing. In the third stage, the binary representation of the network is compressed by using Huffman coding. By using the proposed approach, AlexNet can be compressed from 240MB to 6.9MB, achieving a 35x compression rate, and VGG can be compressed from 552MB to 11.3MB, achieving a 49x compression rate, without loss of accuracy. In [90], an EIE (Efficient Inference Engine) is proposed for processing images while using the compressed network representation. EIE is 189x faster, and 13x faster when compared to CPU and GPU implementations of AlexNet without compression, without loss of accuracy.

Two binary-based network architectures are proposed in [91]: Binary-Weight-Networks and XNOR-Networks. In Binary-Weight-Networks, the filters are approximated with binary values in closed form, resulting in a 32x memory saving. In XNOR-Networks, both the filters and the input to convolutional layers are binary, but non-binary non-linearities like ReLU can still be used. This results in 58x faster convolutional operations on a CPU, by using XNOR-Bitcounting operations. The classification accuracy with a Binary-Weight-Network version of AlexNet is only 2.9% less than the full-precision AlexNet (in top-1 measure),;while XNOR-Networks have a larger, 12.4%, drop in accuracy.

The compression and quantization of DNNs is an area under development. It is expected that easier ways for reducing the size of DNNs will be developed in the near future, and that deep learning frameworks and tools will make them available for use by developers of DNN based applications.

## 3 Designing Deep-learning based Vision Applications

For designers of a deep-learning based vision system, it is important to know which learning frameworks and databases are available to be used in the design process, which DNN models are best suited for their application, whether an already trained DNN can be adapted to the new application, and how to approach the learning process. All these issues are addressed in this section. These guidelines can be used in the design process of any kind of vision system, including robot vision systems.

### 3.1 Developing Tools: Learning Frameworks and Databases

Given the complexity in the design and training of DNNs, special learning frameworks/tools are used for these tasks. There is a wide variety of frameworks available for training and using deep neural networks (see a list in [164]), as well as several public databases suited to various applications and used for training (see a list in [165]). Table 3 presents some of these frameworks, showing the DNN models included in each case. Some of the most popular learning frameworks are described in the following paragraphs. These frameworks were selected by considering the availability of pre-trained models, training/testing speed, and specific advantages such as the inclusion of RBMs in Torch, or multi GPU optimization in CNTK.

Caffe [60] is a deep learning framework written in C++, able to use CUDA [152] and CUDNN [153] for GPU computing. It can also be used on systems without GPUs. Caffe is designed to be used in computer vision applications. The framework can be used through Python and MATLAB, and it has a *model zoo* [178], which is a collection of several state-of-the-art pre-trained models for several computer vision tasks. Pre-trained models from the Caffe Zoo can be downloaded and used in a straightforward way. The model zoo includes AlexNet, CaffeNet (an AlexNet variant used across tutorials), R-CNN, GoogleNet, VGG, and SegNet, among other pre-trained models. Caffe has several dependencies, and its installation process is not straightforward.

Torch [130] is a scientific computing framework with wide support for machine learning. It is written in C, and able to use CUDA [152] and CUDNN [153]. It can be accessed by using a scripting language called LuaJIT. As it is a general machine learning framework, it is able to train LSTMs and energy-based models such as DBMs. It is also able to load some models from the Caffe Model Zoo. It is embeddable, with ports to iOS, Android, and FPGA backends.

Theano [131][132] is a deep learning framework written in Python, that is able to use CUDA [152] and CUDNN [153]. It can run as fast as C frameworks with large datasets and GPUs. It enables the use of symbolic differentiation, which speeds development. It can also generate customized C code for many mathematical operations and its installation procedure is straightforward. There are other frameworks built on Theano, such as Pylearn2 [133], Theano Lights [134], Blocks [135] and Lasagne [136]. Pylearn includes algorithms such as CNNs, LSTMs, and auto-encoders. Theano is able to read Caffe models by using Lasagne.

Tensorflow [137] is an open source library for numerical computation using data flow graphs. Nodes represent operations, while edges represent multidimensional data arrays (tensors). Data flow graphs can be shown using a graphical interface that enables monitoring the system in real-time. It is able to make computations using several CPUs and GPUs, by using CUDA [152] and CUDNN [153]. The graphs can be implemented in Python and C++. Also, Tensorflow is able to generate stand-alone code for Python and C++. Tensorflow is able to load models from Caffe Zoo by installing an extension.

CNTK [141] is the Computational Network Toolkit by Microsoft Research. It describes deep networks as a series of computational steps via directed graphs. Leaf nodes represent inputs, and other nodes represent matrix operations. It enables easy use of feed-forward DNNs, CNNs, and LSTMs. CNTK internal implementation is based on C++. It is able to make computations using several CPUs and GPUs, by using CUDA [152] and CUDNN [153]. CNTK is able to outperform other frameworks in training speed when multiple GPUs are used.

For the designer, it is very important to select the framework best suited to the needs and resources of the final application. In [170] five popular frameworks are compared (Caffe, Neon, Tensorflow, Theano, and Torch). The main conclusions are: (i) Caffe is the easiest to use tool when using standard DNNs, (ii) Theano and Torch are the most extensible tools, (iii) Theano/Torch is fastest for smaller/larger networks for GPU-based training of CNN and fully connected networks; in the case of deployment of CNN and fully connected networks, Torch obtains the best performance followed by Theano, (iv) Theano results in the best performance for GPU-based training and deployment of recurrent networks, and (v) Torch performs the best followed by Theano for CPU-based training and deployment of any DNN.

Table 3. Tools for Designing and Developing deep learning based Vision Applications.

| Tool | Description | Included DNN models |
| --- | --- | --- |
| Caffe [60] | Deep learning framework, written in C++ and able to use CUDA/CuDNN for multi-GPU. Support for Python and MATLAB. | Caffe Model Zoo [3] includes pre-trained reference models of CaffeNet, AlexNet, R-CNN, GoogLeNet, NiN, VGG, Places-CNN, FCN, ParseNet, SegNet, among others. |
| Torch [130] | Scientific computing framework, written in C and able to use CUDA/CuDNN for multi-GPU. Support for LuaJIT. | Torch Model Zoo includes pre-trained models of OverFeat, DeepCompare, and models loaded from Caffe into Torch. |
| Theano [131][132] | Python library for large-scale computing, able to use CUDA/CuDNN. It has started experimental multi-GPU support. | Auto-encoders, RBMs (through Pylearn2), CNN, LSTM (through Theano Lights), models from Caffe Zoo (through Lasagne) |
| TensorFlow [137] | Framework for computation using data flow graphs, written in C++ with Python APIs. Able to use CUDA/CuDNN for multi-GPU/multi-machine. | It has examples of small CNNs and RNN with LSTM in its tutorial. Can load models from Caffe Zoo. |
| MXNet [138] | Deep learning framework. It is portable, allows multi-GPU/multi-machine use, and has bindings for Python, R, C++ and Julia. | Includes three pre-trained models: Inception-BN Network, Inception-V3 Network, and Full ImageNet Network. |

---

[3] Model Zoo refers to a repository of pre-trained models.

| Deeplearning4j [139] | Deep learning library written in Java and Scala. Has multi-GPU/multi-machine support. | Examples of RBM, DBM, LSTM. Its Model Zoo includes AlexNet, LeNet, VGGNetA, VGGNetD. |
|---|---|---|
| Chainer [140] | Deep Learning Framework written in Python. Implements *CuPy* for multi-GPU support. | AlexNet, GoogLeNet, NiN, MLP. Can import some pre-trained models from the Caffe Zoo. |
| CNTK [141] | Computational Network Toolkit. Allows efficient multi-GPU/multi-machine use. | It has no pre-trained models. |
| OverFeat [142] | Feature extractor and classifier based on CNNs. No support for GPUs. | OverFeat Network (pre-trained on ImageNet) |
| SINGA [143][144] | Distributed deep learning platform written in C++. Has support for multi-GPU/multi-machine. | It has no pre-trained models. |
| ConvNetJS [145] | Javascript library for training deep learning models entirely in a web browser. No support for GPUs. | Browser demos of CNNs |
| Cuda-convnet2 [146] | A fast C++/CUDA implementation of convolutional neural networks. Has support for multi-GPU. | It has no pre-trained models. |
| MatConvNet [147] | MATLAB toolbox implementing CNNs. Able to use CUDA/CuDNN on multi-GPU. | Pre-trained models of VGG-Face, FCNs, ResNet, GoogLeNet, VGG-VD [21], VGG-S,M,F [74] CaffeNet, AlexNet. |
| Neon [148] | Python based Deep Learning framework. Able to use CUDA for multi-GPU. | Pre-trained models of VGG, Reinforcement learning, ResNet, Image Captioning, Sentiment analysis, and more. |
| Veles [149] | Distributed platform, able to use CUDA/CuDNN on multi-GPU/multi-machine. Written in Python. | AlexNet, FCNs, CNNs, auto-encoders. |

The availability of training data is also a crucial issue. There are several datasets available for popular computer vision tasks such as image classification, object detection and recognition, semantic segmentation, and action recognition (see Table 4). However, in most of the cases there is no specific database for the particular problem being solved, and pre-training and fine-tuning can be applied (see Section 3.2).

Table 4. Training Databases used in some selected Computer Vision Applications.

| Application | Selected databases |
|---|---|

| Image classification | MNIST (70,000 images, handwritten digits) [171], CIFAR-10 (60,000 tiny images) [172], CIFAR-100 (60,000 tiny images) [172], ImageNet [163] (14M+ images), SUN (scene classification, 130,519 images) [162] |
|---|---|
| Object detection and recognition | Pascal VOC 2012 (11,540 train/val images) [154], KITTI (7,481 train images, car/pedestrian/cyclist) [157], MS COCO (300,000+ images) [158], LabelMe (2,920 train images) [159] |
| Semantic segmentation | Cityscapes (5,000 fine + 20,000 coarse images) [160], Pascal VOC 2012 (2,913 images) [154], NYU2 (RGBD, 1,449 images) [155] |
| Action recognition | MPII (25,000+ images) [156], Pascal VOC 2012 (5,888 images) [154] |

## 3.2 Pre-training and Fine-tuning

When developing a new application, it is very common to use a pre-trained DNN instead of training a network from scratch. The main reason for this is that a large dataset is needed for training a DNN, and usually such a dataset is not available. Pre-training basically consists of using the features of a DNN trained by using a large dataset and applying it to another, usually smaller dataset, after a fine-tuning process. For instance, consider a DNN that has been trained for recognizing a given set of objects using a large dataset containing those objects, and that has to be used for recognizing a different set of objects. The dataset for this second set of objects is smaller. Considering that the two datasets are different, the DNN needs to be fine-tuned. In the case of using a CNN, this problem is normally addressed using a pre-trained CNN, whose last layers (the ones that make the final classification) are replaced and whose first layers are re-used. The rationale behind the procedure is that the same features can be used for solving the two different problems, and that only the classifiers need to be trained again. The training of the new network, including both reused and new layers, is called fine-tuning. The new layers must always be trained by using the second set of objects. Also, the reused layers can be trained further with a smaller learning rate depending on data availability.

An important issue that needs to be analyzed carefully when features of one network are used in a second one is the transferability gap, which grows when the distance between the tasks/problems to be addressed increases [76]. Useful guidelines for the process of using features of one network in a second one can be found in [76].

In the case of robotics applications, pre-training and fine-tuning are relevant since it is difficult to collect large databases, required to train networks, from scratch. Also, the pre-trained networks can be run on the real robotic platforms for estimating the runtime of the corresponding fine-tuned networks. The runtime measured on the robotic platform considers all the restrictions, such as available memory, presence of GPU, CPU power, etc. That measurement can be useful for selecting candidate network architectures without needing to train all the options beforehand.

## 3.3. Selection of the DNN Model

The selection of the network model to be used must consider a trade-off between classification performance, computational requirements, and processing speed.

When selecting a DNN model, there are two main options: using a pre-existing model, or using a novel one. But, the use of a fully novel DNN model is not recommended because of the difficulties in predicting its behavior, unless the purpose of the developer is to propose a new optimized neural model. When using a pre-existing model there are three options: (i) using a pre-trained neural model for solving the task directly, (ii) fine-tuning the parameters of a pre-trained model for adapting it to the new task, or (iii) training the model from scratch. In cases (ii) and (iii), the dimensionality of the network output can be different from the dimensionality required by the task. In that case, the last fully-connected layers of the network in charge of the classification can be replaced by new ones, or by statistical classifiers such as SVMs or random forests.

Table 5 shows popular DNNs, the number of layers and parameters used in each case, the main applications in which the DNNs have been used, and the sizes of the datasets that have been used for training. Table 2 shows the best performing DNNs for each task/application.

It is important to stress that in robotics applications, hardware capabilities like memory and availability of GPU must be considered when selecting the DNN model, since in many cases the runtime requirements of the best performing models cannot be satisfied. Therefore, in cases where the best available models shown in Table 2 cannot be run on the available hardware platforms, or when the runtime requirements cannot be fulfilled, smaller DNN architectures must be considered. For instance, in [198] two smaller CNN based robot detectors that are able to run in real-time, in NAO robots while playing soccer, were presented. Each detector is able to process a robot object-proposal in ~1ms, with an average number of 1.5 proposals per frame. The obtained detection rate was ~97%.

Table 5. Popular CNN-based Architectures used in Computer-vision Applications. The dataset size does not consider data augmentation.

| Name | Layers | Parameters | Application | Dataset size used for training (# images) |
|---|---|---|---|---|
| Le-Net5 | 2 conv, 2 fc, 1 Gaussian | 60 K | Image classification | 70 K |
| AlexNet [20] | 5 conv, 3 fc | 60 M | Image classification | 1.2 M |
| VGG-Fast/Medium/Slow | 5 conv, 3 fc | 77M / 102M / 102M | Image classification | 1.2 M |
| VGG16 | 13 conv, 3 fc | 138 M | Image classification | 1.2 M |
| VGG19 | 16 conv, 3 fc | 144 M | Image classification | 1.2 M |
| GoogleNet | 22 layers | 7 M | Image classification | 1.2 M |
| ResNet-50 [23] | 49 res-conv + 1 fc | 0.75 M[4] | Image classification (also used in other applications) | 1.2 M |
| DenseNet [235] | 1 conv + 1 pooling + N | 0.8M – 27.2M | Image classification | 1.2 M |

---

[4] Trained on CIFAR small-image dataset.

| | | | | |
|---|---|---|---|---|
| | dense blocks + 1 fc | | | |
| SENet [221] | 1 conv + 1 pooling + N squeeze-and-excitation blocks + 1 fc | 103MB – 137MB | Image Classification | 1.2 M |
| Faster R-CNN (ZF) [29] | 5 conv shared, 2 conv RPN, 1 fc reg, 1 fc cls | 54 M | Object detection and recognition | 200 K |
| Faster R-CNN (VGG16) [29] | 13 conv shared, 2 conv RPN, 1 fc reg, 1 fc cls | 138 M | Object detection and recognition | 200 K |
| Faster R-CNN (ResNet-50) [23] | 49 res-conv shared, 2 conv RPN, 1 fc reg, 1 fc cls | 0.75 M[4] | Object detection and recognition | 200 K |
| SegNet [31] | 13 conv encoder, 13 conv decoder | 29.45 M | Semantic segmentation | 200 K |
| Adelaide_context [166] | VGG-16 based, unary and pairwise nets | Not available | Semantic segmentation | 200 K |
| DeepLabv3 [208] | ResNet based, atrous convolution | Not available | Semantic segmentation | 200 K |
| Pyramid Scene Parsing network [209] | Pyramid pooling modules | 188 MB | Semantic segmentation | 200K |
| Stacked hourglass networks [167] | 42 layers, multiscale skipping connections | 166 MB | Action Recognition | 24 K |
| R*CNN [168] | 5 conv shared, 2 fc for main object, 2 fc for secondary object | 500 MB | Action Recognition | 24 K |

## 3.4 Training

*Training, validation and test dataset definitions.* The available dataset must be divided into three non-overlapping groups: training, validation, and test. The training dataset is used for learning the model parameters (weights) of the network. The validation dataset is used for selecting the hyper-parameters by evaluating the performance of the network under different hyper-parameter configurations. The test dataset is used for characterizing the trained network by estimating its generalization error.

*Data augmentation* is a mechanism used to increase the size of the training datasets by applying some transformations/perturbations to the data in order to generate additional examples. Data augmentation has been used in deep learning (reported, e.g., in [20][74]), but also in traditional object learning methods (e.g. [75]). The most frequently used transformations are geometric ones such as in-plane rotations, flipping, random cropping, and photometric transformations, such as RGB jittering [16], color casting, vignetting, and lens distortion [9]. As shown in [74], the use of data augmentation can increase the performance by about 3%.

*Weights Initialization.* When a network is trained from scratch, a set of initial weights must be specified. Initialization of neural networks is an important topic because, if the initialization is not adequate, the network can converge to bad local minima. Initial weights need to have diversity for breaking symmetries: if all the initial weights are the same, they will converge to the same value per layer, and will not be able to represent the input-output relation. Also, when using sigmoid functions, weights with high values can cause saturation on the neuron's outputs, generating very small gradients that are not useful for learning.

A common approach is to use a uniform distribution for weight initialization. For example, a normalized initialization is proposed in [6]:

$$W \sim U\left[-\frac{\sqrt{6}}{\sqrt{n_{in}+n_{out}}}, \frac{\sqrt{6}}{\sqrt{n_{in}+n_{out}}}\right] \quad (2)$$

where $n_{in}$ is the number of inputs for the neuron, and $n_{out}$ is the number of neurons in the next layer connected to the current one.

For the particular case of networks based on ReLU activation function, the following initialization procedure is proposed:

$$W \sim U\left[-\sqrt{2/n_{in}}, \sqrt{2/n_{in}}\right] \quad (3)$$

Also, when using ReLU units, a small bias near 0.01 can be added to account for the asymmetry of this activation function.

*Hyper-parameters for the learning process.* Once the architecture and initial weights are selected, a tuning of the network to the task dataset must be performed. The learning procedure (gradient descent with momentum) is controlled by several hyper-parameters. Recommended hyperparameters for CNNs [20][74] are: momentum 0.9, weight decay $5*10^{-4}$, and initial learning rate $10^{-2}$. Learning rate is decreased by a factor of 10 when the validation error stops decreasing.

Learning is computationally expensive when compared to final DNN use, and requires large amounts of time, memory, and computing power. In most of the problems, learning can be performed offline in high-end computers, and robotic platforms can be used only for collecting data. Systems based on reinforcement learning, that require interaction of an agent with its environment, have been applied in simulated environments like computer games [190], in which runtime of the full system can be controlled and several trials can be performed in parallel. Furthermore, systems able to learn from interaction with their environment have been developed for some specific tasks like learning eye-hand coordination [48], in which several grippers were used for collecting data, large computer capabilities were available, and supervised learning was adapted for the task. However, online learning on limited hardware platforms is a problem that is far from being solved technically, and is an open research area (see Section 5).

Finally, it is important to stress that the final performance of a DNN depends largely on the implementation details [74], such as the selection of the hyper parameters, fine-tuning, data augmentation, etc. Therefore, this aspect needs to be addressed properly.

## 4 Deep Neural Networks in Robot Vision

Deep learning has already attracted the attention of the robot vision community, and during the last couple of years studies addressing the use of deep learning in robots have been published in robotics conferences. Table 6 shows some of the recently published papers presented at flagship robotics conferences and symposia. The papers have been categorized in four application areas: *Object Detection and Categorization*, *Object Grasping and Manipulation*, *Scene Representation and Classification*, and *Spatiotemporal Vision*. The most representative work and the observed tendencies related to the use of deep learning for addressing robot vision problems in each area are described in the next sections. It is important to note that much of the work described has not been validated in real robots operating under real-world conditions, but only in databases corresponding to robot applications. Therefore, in these cases it is not clear whether or not the proposed methods can work in real-time using a real robot platform.

Table 6. Selected Papers on Robot Vision Applications based on DNN.

| Paper | DNN Model and Distinctive Methods | Application |
|---|---|---|
| Pasquale et al., 2015 [38] | Use of CaffeNet in humanoid robots for recognizing objects. | Object Detection and Categorization |
| Bogun et al., 2015 [39] | DNN with LSTM for recognizing objects in videos. | Object Detection and Categorization |
| Hosang at al., 2015 [40] | AlexNet-based R-CNN for pedestrian detection with SquaresChnFtrs as person proposal. | Object Detection and Categorization (Pedestrian detection) |
| Tome et al., 2016 [42] | CNN (Alexnet v/s GoogleNet) for pedestrian detection with LCDF as person proposal. | Object Detection and Categorization (Pedestrian detection) |
| Lenz at al., 2013 [45] | CNN trained on hand-labeled data, two-stage with two hidden layers each. | Object Grasping and Manipulation |
| Redmon et al., 2015 [46] | CNN based on AlexNet trained on hand-labeled data, rectangle regression. | Object Grasping and Manipulation |
| Sung et al., 2016 [47] | Transfer manipulation strategy in embedding space by using a deep neural network. | Object Grasping and Manipulation |
| Levine at al., 2016 [48] | Learn hand-eye coordination independently of camera calibration or robot pose, visual/motor DNN. | Object Grasping and Manipulation |
| Zhou 2014 et al., [50] | CNNs for place recognition based on CaffeNet. | Scene Representation and Classification (Place recognition) |
| Gomez-Ojeda et al., 2015 [49] | CNN for appearance-invariant place recognition, based on CaffeNet. | Scene Representation and Classification (Place recognition) |

| Hou et al., 2015 [51] | CNN for loop closing, based on CaffeNet. | Scene Representation and Classification (Place recognition) |
|---|---|---|
| Sundehauf et al., 2015 [52] | Place categorization and semantic mapping, based on CaffeNet. | Scene Representation and Classification (Scene categorization) |
| Ye et al., 2016 [53] | R-CNN for functional scene understanding, based on Selective Search and VGG. | Scene Representation and Classification (Scene categorization) |
| Cadena et al., 2016 [97] | Multi-modal auto-encoders for semantic segmentation. The inputs are RGB-D, LIDAR and stereo data. It uses inverse depth parametrization. | Scene Representation and Classification (Semantic segmentation, Scene Depth Estimation) |
| Li et al., 2016 [98] | FCN for vehicle detection. Input is a point map generated using a LIDAR. | Object Detection and Categorization |
| Alcantarilla et al., 2016 [99] | Deconvolutional Networks for Street-View Change Detection. | Scene Representation and Classification (Street-View Change Detection) |
| Sunderhauf et al., 2015 [100] | R-CNN used for creating region landmarks for describing an image. AlexNet (up to conv3) as feature extractor. | Scene Representation and Classification (Place Recognition) |
| Albani et al., 2016 [101] | CNN as a validating step for humanoid robot detection. | Object Detection and Categorization (humanoid soccer robot detection) |
| Speck et al., 2016 [102] | CNNs for ball localization in robotics soccer. | Object Detection and Categorization (humanoid soccer ball detection) |
| Finn et al., 2016 [103] | Deep Spatial Auto-encoder for learning state representation. Used for reinforcement learning. | Object Grasping and Manipulation |
| Gao et al., 2016 [104] | Visual CNN and Haptic CNN combined for haptic classification. | Spatiotemporal Vision (Object Understanding) |
| Husain et al., 2016 [105] | Temporal concatenation of the output of pre-trained VGG-16 into a 3D convolutional layer. | Spatiotemporal Vision (Action Recognition) |
| Oliveira et al., 2016 [106] | FCN-based architecture for human body part segmentation. | Object Detection and Categorization |
| Zaki et al. 2016 [107] | Convolutional Hypercube Pyramid based on VGG-f as feature extractor for RGBD. | Object Detection and Categorization |
| Mendes et al., 2016 [108] | Network-in-Network converted into FCN for road segmentation. | Scene Representation and Categorization (Semantic Segmentation) |
| Pinto et al., 2016 [109] | AlexNet based architecture to predict grasp location and angle from image patches, based on self-supervision. | Object Grasping and Manipulation |
| Jain et al., 2016 [110] | Fusion RNN based on LSTM units fed by visual features. | Spatiotemporal Vision (Human Action Prediction). |
| Schlosser et al., 2016 [111] | R-CNN for pedestrian detection by using RGB-D from camera and LIDAR. The depth represented using HHA. RGB-D deformable parts model as object proposals. | Object Detection and Recognition (Pedestrian detection) |
| Costante et al., 2016 [112] | CNNs for learning feature representation and frame to frame motion estimation from optical flow. | Spatiotemporal Vision (Visual Odometry) |
| Liao et al., 2016 [113] | AlexNet-based scene classifier with a semantic segmentation branch. | Scene Representation and Classification (Place Classification / Semantic Segmentation) |
| Mahler et al., 2016 [114] | Multi-View Convolutional Neural Networks with Pre-trained AlexNet as CNNs for computing shape descriptors for 3D objects. Used for selecting object grasps. | Object Grasping and Manipulation |
| Guo et al., 2016 [115] | AlexNet-based CNN for detecting object and grasp by regression on an image patch. | Object Grasping and Manipulation |
| Yin et al., 2016 [116] | Deep Autoencoder for nonlinear time alignment of human skeleton representations. | Spatiotemporal vision (Human Action Recognition) |
| Giusti et al., 2016 [117] | CNN that receives a trail image as input and classifies it as the kind of motion needed for remaining on the trail. | Scene Representation and Classification (Trail Direction Classification) |
| Held et al., 2016 [118] | CaffeNet, pre-trained on ImageNet, fine-tuned for viewpoint invariance. | Object Detection and Categorization (Single-view Object Recognition) |
| Yang et al., 2016 [119] | FCN and DSN based Network with AlexNet as basis. Used with CRF for estimating 3D scene layout from monocular camera. | Scene Representation and Classification (Semantic Segmentation, Scene Depth Estimation) |
| Uršic et al., 2016 [120] | R-CNN used for generating histograms of part-based models. Places-CNN (CaffeNet trained on Places 205) as region feature extractor. | Scene Representation and Classification (Place Classification) |
| Bunel et al., 2016 [121] | CNN architecture of 4 convolutional layers with PReLU as activation function and 2 FC layers. Used for detecting pedestrians at far distance. | Object Detection and Categorization (Pedestrian Detection) |
| Murali et al., 2016 [122] | VGG architecture as feature extractor for unsupervised segmentation of image sequences. | Spatiotemporal Vision (Segmentation of trajectories in robot-assisted surgery). |

| Kendall et al., 2016 [123] | Bayesian PoseNet (Modified GoogLeNet) with pose regression, use dropout for estimating uncertainty. | Scene Representation and Classification (Camera re-localization) |
|---|---|---|
| Husain et al., 2016 [124] | CNN using layers from OverFeat Network with multiple pooling sizes. RGB-D inputs. Use of HHA and distance-from-wall for depth. | Scene Representation and Classification (Semantic Segmentation) |
| Hoffman et al., 2016 [125] | R-CNN using RGB-D data, based on AlexNet and HHA. Proposals are based on RGB-D Selective Search. | Object Detection and Categorization |
| Sunderhauf et al., 2016 [126] | Places-CNN as image classifier for building 2D grid semantic maps. LIDAR used for SLAM. Bayesian filtering over class labels. | Scene Representation and Classification (Place Classification) |
| Saxena et al., 2017 [189] | CNN used for image-based visual servoing. Inputs are monocular images from current and desired poses. Outputs are velocity commands for reaching the desired pose. | Object Grasping and Manipulation (Visual servoing) |
| Lei et al., [191] | Robot exploration by using a CNN, trained first by supervised learning, later by using deep reinforcement learning. Tested on simulated and real experiments. | Scene Representation and Classification (Scene Exploration) |
| Zhu et al., [192] | Robot navigation by learning a scene-dependent Siamese network, which receives images from two places as input, and generates motion commands for travelling between them. | Scene Representation and Classification (Visual Navigation) |
| Mirowski et al., [193] | Robot navigation in complex maze-like environments from raw monocular images and inertial information. Use of deep reinforcement learning on a network composed of a CNN followed by two LSTM layers. Multi-task loss considering reward prediction and depth prediction improves learning. | Scene Representation and Classification (Visual Navigation) |

## 4.1 Object Detection and Categorization

Object detection and categorization is a fundamental ability in robotics. It enables a robot to execute tasks that require interaction with object instances in the real-world. Deep learning is already being used for general-purpose object detection and categorization, as well as for pedestrian detection, and for detecting objects in robotics soccer. State-of-the-art methods used for object detection and categorization are based on generating object proposals, and then classifying them using a DNN, enabling systems to detect thousands of different object categories. As will be shown, one of the main challenges for the application of DNNs for object detection and characterization in robotics is real-time operation. It must be stressed that obtaining the required object proposals for feeding the DNNs is not real-time in the general case, and that, on the other hand, general-purpose object detection and categorization DNN based methods are not able to run in real-time in most robotics platforms. These challenges are addressed by using task-dependent methods for generating few, fast, and high quality proposals for a limited number of possible object categories. These methods are based on using other information sources for segmenting the objects (depth information, motion, color, etc.), and/or by using object specific, non general-purpose, weak detectors, for generating the required proposals. Also, smaller DNN architectures can be used when dealing with a limited number of object categories.

It is worth mentioning that most of the reported studies do not indicate the frame rate needed for full object detection/categorization, or they show frame rates that are far from being real-time. In generic object detection methods, computation of proposals using methods like Selective Search or EdgeBoxes takes most of the time [29]. Systems like Faster R-CNN that compute proposals using CNNs are able to obtain higher frame rates, but require the use of high-end GPUs for working. The use of task-specific knowledge based detectors on depth [33][34][111], motion [38], color segmentation [101], or weak object detectors [42] can be useful for generating fewer, faster proposals, which is the key for achieving high frame rates on CPU-based systems. Finally, methods based on FCNs cannot achieve high frame rates on robotic platforms with low processing capabilities (no GPU available) because they process images with larger resolutions than normal CNNs. Then, FCNs cannot be used trivially for real time robotics on these kinds of platforms.

First, we will analyze the use of complementary information sources for segmenting the objects. Robotic platforms usually have sensors, such as Kinects/Asus or LIDAR sensors, that are able to extract depth information. Depth information can be used for boosting the object segmentation. Methods that use RGBD data for detecting objects include those presented in [33][34][92][107][125], while methods that use LIDAR data include those of [98][111]. These methods are able to generate proposals by using tabletop segmentation or edge/gradient information. Also, they generate colorized images from depth for use in a standard CNN pipeline for object recognition. These studies use CNNs and custom depth-based proposals, and then their speed is limited by the CNN model. For instance, in [197], a system is described that runs at 405 fps on a Titan X GPU, 67 fps on a Tegra X1 GPU, and 62 fps on a Core i7 6700K CPU [197]. It must be noted that, while TitanX GPU and Core i7 CPU processors are designed for desktop computers, Tegra X1 is a mobile GPU processor for embedded systems aimed at low power consumption, and so it can be used on robotic platforms.

Second, we will present two applications that use specific detectors for generating the object proposals: pedestrian detection, and detection of objects in robotics soccer. Detecting pedestrians in real time with high reliability is an important ability in many robotics applications, especially for autonomous driving. Large differences in illumination and variable, cluttered backgrounds outdoors are hard issues to be addressed. Methods that use CNN-based methods for pedestrian detection include [40], [41], [42], [121], [199], [200], [201], [202], [203]. Person detectors, such as LDCF [43] or ACF [44], are used for generating pedestrian proposals in some methods. As indicated in [42], the system based on AlexNet [20] requires only 3 ms for processing a region, and proposal computation runs at 2 fps when using LDCF, and at 21.7 fps when using ACF in an NVIDIA GTX980 GPU, with images of size 640x480. The lightweight pedestrian detector is able to run at 2.4 fps in an NVIDIA Jetson TK1, and beats classical methods by a large margin [42]. A second approach is to use a FCN for detecting parts of persons [106]. This method is tested on a NVIDIA Titan X GPU, and is able to run at 4 fps when processing images of size 300x300. KITTY pedestrians is a benchmark used by state-of-the-art methods. Current leading methods whose algorithms are described (many of the best performing methods do not describe their algorithms) are shown in Table 7. Note that, as the objective of the benchmark is to evaluate accuracy, most of the methods cannot run in real-time (~30 fps). This is an aspect that needs further research.

Table 7. Selected Methods for Pedestrian Detection from KITTY Benchmark. Note that methods not describing their algorithms are not included.

| Method | Moderate | Easy | Hard | Runtime | Computing Environment |
|---|---|---|---|---|---|
| RRC [199] (code available) | 75.33% | 84.14 % | 70.39 % | 3.6 s | GPU @ 2.5 Ghz (Python + C/C++) |
| MS-CNN [200] (code available) | 73.62% | 83.70% | 68.28% | 0.4 s | GPU @ 2.5 Ghz (C/C++) |
| SubCNN [201] | 71.34 % | 83.17 % | 66.36 % | 2 s | GPU @ 3.5 Ghz (Python + C/C++) |
| IVA [202] (code available) | 70.63 % | 83.03 % | 64.68 % | 0.4 s | GPU @ 2.5 Ghz (C/C++) |
| SDP+RPN [203] | 70.20 % | 79.98 % | 64.84 % | 0.4 s | GPU @ 2.5 Ghz (Python + C/C++) |

Robotics soccer is an interesting area because it requires real-time vision algorithms able to work with very limited computational resources. Deep learning techniques developed in robot soccer can therefore be transferred to other resource-constrained platforms, such as smartphones and UAVs. Current applications for robotics soccer include soccer player detection [101][198], and ball detection [102]. State-of-the-art soccer player detection [198] uses a model based on SqueezeNet using only 2 convolutional layers, an extended fire layer and a fully-connected layer, while processing regions of size 28x28. This method requires only 1 ms on an Nao robot for processing a region, using only an Intel Atom CPU. Another application is ball detection, which is addressed in [102]. This method uses an architecture based on 3 conv layers and 2 fully-connected layers. However, its runtime inside a Nao robot is not reported. Currently, CNNs have been used for solving very specific tasks in robotics soccer successfully, and it is expected that new algorithms (and hardware) will enable the use of deep learning in more global, high-level tasks.

### 4.2. Object Grasping and Manipulation

The ability to grasp objects is of paramount importance for autonomous robots. Classical algorithms for grasping and manipulating objects require detecting both the gripper and the object in the image, and using graph-based or Jacobian-based algorithms for computing the movements for the gripper. Deep learning has generated a paradigm change since tasks like grasping, manipulation and visual servoing can be solved directly, without the need of hand-crafting intermediate processes like detecting features, detecting objects, detecting the gripper, or performing graph-based optimizations. As problems are solved as a whole from raw data, the performance limit imposed by inaccuracies of intermediate processes can be surpassed, and, therefore, novel approaches for solving tasks related to grasping, manipulation, and visual servoing are being developed. Also, the new algorithms are able to be generalized to novel objects and situations. The performance of this new generation of methods is currently limited by the size of the training datasets and the computing capabilities, which becomes relevant when using hardware-limited robotic platforms.

Grasping objects directly, without the need of detecting specific objects, is a difficult problem because it must generalize over different kinds of objects, including those not available when training the system. This problem has recently been addressed successfully by using deep learning. Studies that are based on selecting grasp locations include [45], [46], [115] and [109], which uses self-supervision during 700 hours of trial and error in a Baxter robot (see Figure 4). That work has been extended for learning eye-hand coordination in [48], which is the state-of-the-art. The latter system is trained without supervision by using over 800,000 grasp attempts collected over the course of two months, using between 6 and 14 manipulators at any given time. The architecture is based on a concatenation of two inputs: an image (size 512x512), and a motor command. The image network has 7 convolutional layers, and its output is concatenated with the motor command. The concatenation layer is further processed by an additional 8 convolutional layers and two fully-connected layers. The system achieves effective real-time control; the robot can grasp novel objects successfully and correct mistakes by continuous servoing, and lowers the failure rate from 35% (hand designed) to 20%. However, learned eye-hand coordination depends on the particular robot used for training the system. The full computing hardware used in this work is not fully described. It must be noted that these kinds of solutions are able to run successfully in real-time using GPUs.

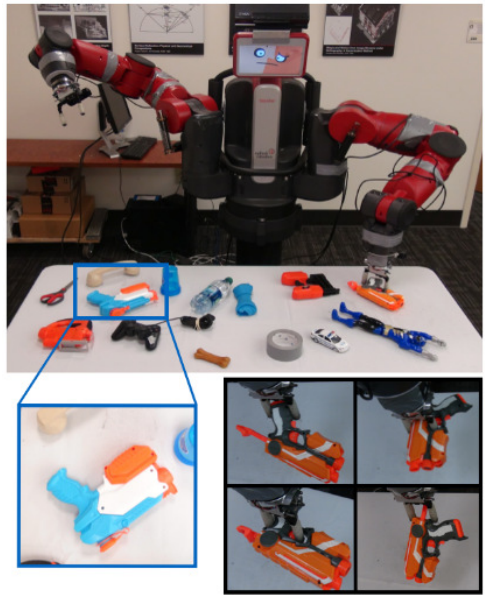

Figure 4. A Baxter Robot learning to Grasp Objects by Trial and Error, by using Self-supervised Learning. Figure taken from [109] with permission of the authors.

Systems able to infer manipulation trajectories for object-task pairs using DNNs have been developed recently. This task is difficult because generalization over different objects and manipulators is needed. They include Robobarista [47], and [103]. These systems are able to learn from manipulation demonstrations and generalize over new objects. In Robobarista [47], the input of the system is composed of a point cloud, a trajectory, and a natural language space, which is mapped to a joint space. Once trained, the trajectory generating the best similarity to a new point cloud and natural language instruction is selected. Computational hardware (GPU) and runtime are not reported in this work. In [103] the system is able to learn new tasks by using image-based reinforcement learning. That system is composed by an encoder and a decoder. The encoder is formed by 3 convolutional layers and 2 fully-connected layers. Computational hardware and runtime are not reported in this work.

Visual servoing systems have also benefited from the use of deep learning techniques. In Saxena et al. [189], a system that is able to perform image-based visual servoing by feeding raw data into convolutional networks is presented. The system does not require 3D geometry of the scene or intrinsic parameters from the camera. The CNN takes a pair of monocular images as input, representing the current and desired pose, and processes them by using 4 convolutional layers and a fully-connected layer. The system is trained for computing the transform in 3D space (position and orientation) needed for moving the camera from its current pose to the desired one. The estimated relative motion is used for generating linear and angular speed commands for moving the camera to the desired pose. The system is trained on synthetic data, and tested both in simulated environments and using a real quadrotor. The system takes 20 msec for processing each frame when using an NVIDIA Titan X GPU, and 65 msec when using a Quadro M2000 GPU. A wifi connection was used for transferring data between the drone and a host PC.

Multimodal data delivers valuable information for object grasping. In [104], a haptic object classifier that is fed by visual and haptic data is proposed. The outputs of a visual CNN and a haptic CNN are fused into a fusion layer, followed by a loss function for haptic adjectives. The visual CNN is based on the (5a) layer of GoogleNet, while the haptic CNN is based on 3 one-dimensional convolutional layers followed by a fully-connected layer. Examples of haptic adjectives are "absorbent," "bumpy," and "compressible". The system, after being trained, is able to estimate haptic adjectives of objects by using only visual data, and is shown to improve over systems based on classical features. Computing hardware and runtime of the method is not reported.

In summary, deep learning has enabled the generation of novel approaches to solving tasks related to object grasping, object manipulation, and visual servoing. Real-time processing is needed for hand-eye coordination and visual servoing to work, which imposes constraints on hardware capabilities. However, real-time processing is not needed for open-loop object grasping detection. Direct methods that do not require detecting objects to work are robust, but also dependent on the robot used for collecting data. GPUs are needed for deep learning based methods that require closed loop control, which limits its usefulness when working with hardware-limited robotic platforms. Grasping and manipulation algorithms will continue to benefit from the use of deep reinforcement learning methods in the near future.

### 4.3 Scene Representation and Classification

Scene representation is an important topic since it is required by mobile robots for performing a variety of tasks that depend on semantic and/or geometric information from the current scene. In particular, visual navigation is an important ability that is needed by autonomous robots for performing more general tasks that involve moving autonomously. Classical approaches for representing scenes from monocular images are based on extraction of hand-crafted features, while visual navigation requires the robot to have previously had a metric or topological map of its environment as well as an estimation of the depth of the obstacles/objects in the current frame. Deep learning has enabled solving a variety of scene-related tasks directly, ranging from encoding scene information to performing navigation from raw monocular data, without the need of handcrafting sub-processes like feature extraction, map estimation, or localization. These novel task-oriented algorithms are able to solve difficult problems successfully that were not directly affordable before. In the following paragraphs the use of deep learning in applications such as place recognition, semantic segmentation, visual navigation, and 3D geometry estimation is analyzed.

Place recognition is a task that has been solved successfully by using deep learning-based approaches. Studies related to this problem are [49], [50], [51], [52], [113], [120] and [100]. These methods are able to recognize places under severe appearance changes caused by weather, illumination, and changes in viewpoint. Also, in [126], data from LIDAR is used alongside a CNN for building a tridimensional semantic map. The CNN used is based on AlexNet and requires 30 ms to run on an Nvidia Quadro K4000 GPU. There are several datasets for testing algorithms, such as Sun [204], and Places [50][205], and also several variants derived from them. As there is not a dominant benchmark for place recognition, different studies use different datasets, sometimes customized, and thus most methodologies cannot be compared directly. Also, the best performing methods in ILSVRC, scene classification challenge [216], and Scene2 challenges [217], do not describe their algorithms in detail, but achieve an impressive 0.0901 top-5 classification error. Computing hardware and runtime related to challenges [216] and [217] are not reported.

Nevertheless, the related datasets used in scene classification, and the best performing methods in each case, are described in Table 8.

Table 8 Datasets used for Scene Classification.

| Dataset | Results |
| --- | --- |
| Places [50] | 50.0% [50] (CaffeNet), 14.56 ms @ Pascal Titan X<br>85.16% [113] (AlexNet for scene classification, 8 conv for semantic segmentation) |
| Sun [204][206] | 38% [204] (non deep)<br>66.2% [50] (AlexNet), 14.56 ms @ Pascal Titan X |
| Sun-RGBD[113] | 41.3% [113] (input 81x81, 5 conv + 3 fc classification, 8 conv semantic segmentation) |

Semantic segmentation consists of classifying each pixel in the image, enabling the solving of tasks that require a pixel-wise level of precision. Studies related to semantic segmentation of scenes are [98], [108], [113], [208], [209], [210], [211], [212], [213], [214] and [215]. Also, multimodal information is used in [97], [124] and [126]. These studies are useful for tasks that require pixel-wise precision like road segmentation and other tasks in autonomous driving. However, runtimes of methods based on semantic segmentation are usually not reported. PASCAL VOC and Cityscapes are two benchmarks used for semantic segmentation. While images from PASCAL VOC are general-purpose, Cityscapes is aimed at autonomous driving. Best performing methods in both databases are reported in Tables 9 and 10. It can be noted that DeepLabv3 [208] and PSPNet [209] achieve good average precision at both benchmarks, and also code from [209] is available. Thus, [209] is recommended. Also, the increasing performance in average precision has a deep impact on autonomous driving, as this application requires confident segmentation for working because of the risks involved in it.

Table 9: PASCAL VOC Segmentation Benchmark. Only entries with reported methods are considered. Repeated entries are not considered.

| Name | AP | Runtime (msec/frame) |
| --- | --- | --- |
| DeepLabv3-JFT [208] | 86.9 | n/a |
| DIS [213] | 86.8 | 140 msec (gpu model not reported) |
| CASIA_IVA_SDN [214] | 86.6 | n/a |
| IDW-CNN [215] | 86.3 | n/a |
| PSPNet [209] (code available) | 85.3 | n/a |

Table 10: Cityscapes Benchmark. Only entries with reported methods are considered. Repeated entries are not considered.

| Name | IoU class | Runtime (ms/frame) |
| --- | --- | --- |
| Deeplabv3 [208] | 81.3 | n/a |
| PSPNet [209] | 81.2 | n/a |
| ResNet-38 [210] | 80.6 | 6 msec (minibatch size 10) @ GTX 980 gpu |
| TuSimple_Coarse [211] (code available) | 80.1 | n/a |
| SAC-multiple [212] | 78.2 | n/a |

Visual navigation using deep learning techniques is an active research topic since new methodologies are able to solve navigation directly, without the need of detecting objects or roads. Examples of studies related to visual navigation are [117], [93], [191], [192] and [193]. Their methods are able to perform navigation directly over raw images captured by RGB cameras. For instance, the forest trail follower in [117] has an architecture composed of 7 convolutional layers and 1 fully-connected layer. It is able to run at 15 fps on an Odroid-U3 platform, enabling it to run in a real UAV.

The use of CNN methodologies for estimating 3D geometry is another active research topic in robotics. Studies related to 3D geometry estimation from RGB images include [112], [119], [123], Also, depth information is added in [97] and [124]. These methods provide a new approach to dealing with 3D geometry, and are shown to overcome the classical structure-from-motion approaches, as they are able to infer depth even from only a single image.

A functional area is a zone in the real-world that can be manipulated by a robot. Functional areas are classified by the kind of manipulation action the robot can realize on them. In [53] a system for localizing and recognizing functional areas by using deep learning is presented. The system enables an autonomous robot to have a functional understanding of an arbitrary indoor scene. An ontology considering 11 possible end categories is set. Functional areas are detected by using selective-search region proposals, and then by applying a VGG-based CNN trained on the functionality categories. The system is able to generalize onto novel scenes. Neither hardware nor runtime are reported. However, its architecture is similar to RCNN [27], which requires 18 seconds for processing a frame on a K20x GPU.

In summary, deep learning has given rise to a new generation of task-oriented algorithms that do not require explicit subtask engineering. In general, methods for representing, recognizing, or classifying scenes as a whole (not pixel-wise) do not require computing object proposals; the image is processed directly by a CNN, and then, depending on the computing capabilities of the robotic platform, near real time processing can be achieved. However, pixel-wise methods based on FCNs (like semantic segmentation) require a GPU for running in near real time, because a larger resolution is needed by the intermediate convolutional layers. When using robotic platforms with low computing capabilities (only CPU), processing each frame in an FCN can take several seconds. Considering that CPUs are also needed to execute other tasks concurrently, the use of FCN-based methods on CPUs is discouraged for real time applications in most robotic platforms.

### 4.4 Spatiotemporal Vision

The processing of video and other spatiotemporal sources of information is of paramount importance in robotics. Classical methods related to spatiotemporal vision require engineering subtasks such as motion detection, optical flow, local feature extraction, or image segmentation. Deep learning has the potential to change the current paradigm of spatiotemporal vision from hand-engineering complex subtasks to solving tasks as a whole. The increment in both video datasets and computational power will enable the analysis of spatiotemporal data at near real-time, but only in the future. Then, spatiotemporal vision will be a goal of future robot vision as well as computer vision research.

The use of short video sequences for object recognition is an alternative for improving current image-based techniques. In [39], a convolutional LSTM is proposed for processing short video sequences, those containing about four frames, for robotic perception. The system is shown to improve on the baseline (smoothing CNN results from individual image frames), and is able to run at 0.87 fps when using a GPU.

Human action recognition is an area that can benefit greatly from using video sequences instead of isolated images. Reports of research dealing with recognition of human actions include [229], [230], [231], [232], [233], [234], [105], [116], and [110], in which driver activity anticipation is performed. In various studies, information from multiple frames is integrated by using several CNNs for each frame, by using LSTM or by using Dynamic Time Warping. The C3D method [234] is able to run at 313 fps on a K40 GPU, processing videos with resolution 320x240. Thus, those kinds of methods have the potential for running in real time. However, current state-of-the-art methods are not real time, or their runtimes are not reported. The best performing methods on the UCF-101 dataset are reported in Table 11. Note that both the accuracy and runtime of C3D [234] are lower than Wu et al. [232], and so a tradeoff between accuracy and runtime is present. For selecting a method, it must be considered if real time is needed in the specific task to be solved.

Table 11: Current Results on the UCF-101 Video Dataset for Action Recognition [230]

| Name | Accuracy | Runtime (msec/frame) |
| --- | --- | --- |
| C3D (1 net) + linear SVM [234] | 82.3% | 3.2 msec/frame @ Tesla K40 GPU |
| VGG-3D + C3D [105] | 86.7% | n/a |
| Ng et al. [231] | 88.6% | n/a |
| Wu et al. [232] | 92.6% | 1730 msec/frame @ Tesla K80 GPU |
| Guo et al. [230] | 93.3% | n/a |
| Lev et al. [233] | 94.0% | n/a |

Transfer of video-based computer vision techniques to medical robotics is an active area of research, especially in surgery-related applications. The use of deep learning in this application area has already begun. For instance, in [122], a system that is able to segment video and kinematic data for performing unsupervised trajectory segmentation of multi-modal surgical demonstrations is implemented. The system is able to segment video-kinematic descriptions of surgical demonstrations successfully corresponding to stages such as "position," "push needle," "pull needle," and "hand-off." The system is based on a switching linear dynamical system which considers both continuous and discrete states composed of kinematic and visual features. Visual features are extracted from the fifth convolutional layer of a VGG CNN and then reduced to 100 dimensions by using PCA. Clustering on the state space is applied by the system and it is able to learn transitions between clusters, enabling trajectory segmentation. The system uses a video source with 10 frames per second; however, the frame rate of the full method is not reported.

Spatiotemporal vision applications need real-time processing capability to be useful in robotics. However, current investigations using deep learning are experimental, and the frame rates for most of the methods are not reported. Platforms with low computing capability (only CPU) are not able to run most of the methods at a useful frame rate. Availability of specialized hardware for computing CNNs [12] would enable the use of deep spatiotemporal vision applications in the future.

## 5. Discussion: Current and Future Challenges of Deep Learning in Robotic Applications

As already mentioned, the increase in the use of deep learning in robotics will depend on the possibility that DNN methods will be able to adapt to the requirements of robotics applications, such as real-time operation with limited on-board computational resources, and the ability to deal with constrained observational conditions derived from the robot geometry, limited camera resolution, and sensor/object relative pose.

The current requirements of large computational power and memory needed for most CNN models used in vision applications are relevant barriers to its adoption in resource-constrained robotic applications such as UAVs, humanoid biped robots, or autonomous vehicles. Efforts are already being made to develop CNN architectures that can be compressed and quantized for use in resource-constrained platforms [89][90] (see Section 2.8). For instance, in [198] a CNN based robot detector is presented that is able to run in real-time, in NAO robots while playing soccer. In addition, companies such as Intel, NVIDIA, and Samsung, just to name a few, are developing CNN chips that will enable real-time vision applications [12]. For instance, mobile GPU processors like NVIDIA Tegra K1 enable efficient implementation of deep learning algorithms with low power consumption, which is relevant for mobile robotics [197]. It is expected that these methodologies will consolidate in the next few years, and will then be available to the developers of robot vision applications.

The ability of deep learning models to manage spatial invariance to the input data in a computationally and efficient manner is another important issue. In order to address this, the Spatial Transformer Module (STM) was recently introduced [179]. This corresponds to a self-contained module that can be incorporated into DNNs, and that is able to perform explicit spatial transformation of the features. The most important characteristic of STM is that its parameters (i.e. the transformations of the spatial transformation) can be learned together with the other parameters of the network using the back propagation of the loss. The further development of the STM concept will allow addressing the required invariance to the observational conditions derived from the robot geometry, limited camera resolution, and sensor/object relative pose. STMs are able to deal with rotations, translations, and scaling [179], and STMs are already being extended to deal with 3D transformations [180]. Further development of STM-inspired techniques is expected in the next few years.

Unsupervised learning is another relevant area of future development for deep-learning based vision systems. Biological learning is largely unsupervised; animals and humans discover the structure of the world by exploring and observing it. Therefore, one would expect that similar mechanisms could be used in robotics and other computer-based systems. Until now, the success of purely supervised learning, which is based largely on the availability of massive labeled data, has overshadowed the application of unsupervised learning in the development of vision systems. However, when considering the increasing availability of non-labeled digital data, and the increasing number of vision applications that need to be addressed, one would expect further development of unsupervised learning strategies for DNN models [12][16]. In the case of robotics applications, a natural way of addressing this issue is by using deep learning and reinforcement learning together. This strategy has already been applied for the learning of games (Atari, Go, etc.) [190][196], for solving simple tasks like pushing a box [103], and for solving more complex tasks such as navigating on simulated environments [192][193], but not for robots learning in the wild. The visual system of most animals is active, i.e. animals decide where to look, based on a combination of internal and external stimuli. In a robotics system having a similar active vision approach, reinforcement learning can be used for deciding where to look according to the results of the interaction of the robot with the environment.

A related relevant issue is that of open-world learning [181], i.e. to learn to detect new classes incrementally, or to learn to distinguish among subclasses incrementally after the "main" one has been learned. If this can be done without supervision, new classifiers can be built based on those that already exist, greatly reducing the effort required to learn new object classes. Note that humans are continuously inventing new objects, fashion changes, etc., and therefore robot vision systems will need to be updated continuously, adding new classes, and/or updating existing ones [181]. Some recent work has addressed these issues, based mainly on the joint use of deep-learning and transfer-learning methods [182][183].

Classification using a very large number of classes is another important challenge to address. AlexNet, like most CNN networks, was developed for solving problems which contain ~1,000 classes (e.g., ILSVRC 2012 challenge). It uses fully connected units and a final softmax unit for classification. However, when working on problems with a very large number of classes (10,000+) in which dense data sampling is available, nearest neighbors could work better than other classifiers [188]. Also, the use of a hierarchy-aware cost function (based on WordNet) could enable providing more detailed information when the classes have redundancy [188]. However, in current object recognition benchmarks (like ILSVRC 2016), the number of classes has not increased, and no hierarchical information has been used. The ability of learning systems to deal with a high number of classes is an important challenge that needs to be addressed for performing open-world learning.

Another important area of research is the combination of new deep learning based methods with classical vision methods based on geometry, which has been very successful in robotics (e.g. visual SLAM). This topic has been addressed at recent workshops [58] and conferences [59]. There are several ways in which these important and complementary paradigms can be combined. On the one hand, geometry-based methods such as Structure from Motion and Visual SLAM can be used for the training of deep-learning based vision systems; geometry-based methods can extract and model the structure of the environment, and this structure can be used for assisting the learning process of a DNN. On the other hand, DNNs can also be used for learning to compute the visual odometry automatically, and eventually to learn the parameters of a visual SLAM system. In a very recent study, STMs were extended to 3D spatial transformations [180], allowing end-to-end learning of DNNs for computing visual odometry. It is expected that STM-based techniques will allow end-to-end training for optical flow, depth estimation, place recognition with geometric invariance, small-scale bundle adjustment, etc. [180].

A final relevant research topic is the direct analysis of video sequences in robotics applications. The analysis of video sequences by using current deep-learning based techniques is very demanding of computational resources. It is expected that the increase of computing capabilities will enable the analysis of video sequences in real-time in the near future. Also, the availability of labeled video datasets for learning tasks such as action recognition will enable the improvement of robot perception tasks, which are currently addressed by using independent CNNs for individual video frames.

# 6. Conclusions

In light of the paradigm shift that deep-learning has produced in pattern recognition and machine learning, this survey has addressed the use of DNNs in robot vision. After providing a comprehensive overview of DNNs, which includes presenting the historical development of the discipline, basic definitions, popular DNN models, and application areas in computer vision, the design process of DNN-based vision systems has been presented, followed by a review of the application of deep-learning in robot vision. Research tendencies are shown and the main studies are grouped together in four application areas: Object Detection and Categorization, Object Grasping and Manipulation, Scene Representation and Classification, and Spatiotemporal Vision. Then, a discussion about the future of deep learning in robotic vision has been presented.

We believe that this survey provides a valuable guide for the developers of robot vision systems, since it promotes understanding of the basic concepts behind the application of deep-learning in vision applications, explains the tools and frameworks used in the development process of vision systems, and shows current tendencies in the use of DNN models in robot vision.

It is expected that the use of deep learning in robot vision will increase in the next few years, thanks to a better adaptation of DNN-based solutions to the requirements of robotics applications. Areas in which advances will be observed in the coming years are DNN models with fewer computational and memory requirements, and DNN models that are invariant to different transformations in the input data. Moreover, given that in many applications robots need to learn as they interact with the real-world, more research will be done in unsupervised learning, open-world learning, and in the joint use of deep and reinforcement learning. Finally, it is expected that joint use of geometry-based and deep-based methods will allow developing state-of-the-art vision systems that will allow increasing the autonomy of robotic systems, and also the automatic training of DNNs.

## Acknowledgments

This work was partially funded by FONDECYT grant 1161500.